# Agglomerative Bregman Clustering

**Matus Telgarsky**                                                                    MTELGARS@CS.UCSD.EDU
**Sanjoy Dasgupta**                                                                    DASGUPTA@CS.UCSD.EDU
Department of Computer Science and Engineering, UCSD, 9500 Gilman Drive, La Jolla, CA 92093-0404

## Abstract

This manuscript develops the theory of agglomerative clustering with Bregman divergences. Geometric smoothing techniques are developed to deal with degenerate clusters. To allow for cluster models based on exponential families with overcomplete representations, Bregman divergences are developed for nondifferentiable convex functions.

## 1. Introduction

Starting with points $\{x_i\}_{i=1}^m$ and a pairwise merge cost $\Delta(\cdot,\cdot)$, classical agglomerative clustering produces a single hierarchical tree as follows (Duda et al., 2001).

1. Start with $m$ clusters: $C_i := \{x_i\}$ for each $i$.

2. While at least two clusters remain:
   (a) Choose $\{C_i, C_j\}$ with minimal $\Delta(C_i, C_j)$.
   (b) Remove $\{C_i, C_j\}$, add in $C_i \cup C_j$.

In order to build a hierarchy with low $k$-means cost, one can use the merge cost due to Ward (1963),

$$\Delta_w(C_i, C_j) := \frac{|C_i||C_j|}{|C_i| + |C_j|} \|\tau(C_i) - \tau(C_j)\|_2^2,$$

where $\tau(C)$ denotes the mean of cluster $C$.

The $k$-means cost, and thus the Ward merge rule, inherently prefer spherical clusters of common radius. To accommodate other cluster shapes and input domains, the squared Euclidean norm may be replaced with a relaxation sharing many of the same properties, a *Bregman divergence*.

This manuscript develops the theory of agglomerative clustering with Bregman divergences.

### 1.1. Bregman clustering

There is already a rich theory of clustering with Bregman divergences, and in particular the relationship of these divergences with exponential family distributions (Banerjee et al., 2005). The standard development has two shortcomings, the first of which is amplified in the agglomerative setting.

**Degenerate divergences.** Many divergences lead to merge costs which are undefined on certain inputs. This scenario is exacerbated with small clusters; for instance, with Gaussian clusters, the corresponding divergence rule is the KL divergence, which demands full rank cluster covariances. This is impossible with $\leq d$ points, but the agglomerative procedure above starts with singletons.

**Minimal representations.** The standard theory of exponential families and its connections to Bregman divergences depend on *minimal representations*: there is just one way to write down any particular distribution. On the other hand, the natural encoding for many problems — e.g., Ising models, and many other examples listed in the textbook of Wainwright & Jordan (2008, Section 4) — is *overcomplete*, necessitating potentially tedious conversions to invoke the theory.

### 1.2. Contribution

The approach of this manuscript is to carefully build a theory of Bregman divergences constructed from convex, yet nondifferentiable functions. Section 2 will present the basic definition, and verify this generalization satisfies the usual Bregman divergence properties.

Section 3 will revisit the standard Bregman hard clustering model (Banerjee et al., 2005), and show how it naturally leads to a merge cost $\Delta$. Section 4 then constructs exponential families, demonstrating that nondifferentiable Bregman divergences, while permitting representations which are not minimal, still satisfy all the usual properties. To overcome the aforemen-tioned small-sample cases where divergences may not be well-defined, Section 5 presents smoothing procedures which immediately follow from the preceding technical development.

To close, Section 6 presents the final algorithm, and Section 7 provides experimental validation both by measuring cluster fit, and the suitability of cluster features in supervised learning tasks.

The various appendices contain all proofs, as well as some additional technical material and examples.

### 1.3. Related work

A number of works present agglomerative schemes for clustering with exponential families, from the perspective of KL divergences between distributions, or the analogous goal of maximizing model likelihood, or lastly in connection to the information bottleneck method (Iwayama & Tokunaga, 1995; Fraley, 1998; Heller & Ghahramani, 2005; Garcia et al., 2010; Blundell et al., 2010; Slonim & Tishby, 1999). Furthermore, Merugu (2006) studied the same algorithm as the present work, phrased in terms of Bregman divergences. These preceding methods either do not explicitly mention divergence degeneracies, or circumvent them with Bayesian techniques, a connection discussed in Section 5.

Bregman divergences for nondifferentiable functions have been studied in a number of places. Remark 2.4 shows the relationship between the presented version, and one due to Gordon (1999). Furthermore, Kiwiel (1995) presents divergences almost identical to those here, but the manuscripts and focuses differ thereafter.

The development here of exponential families and related Bregman properties generalizes results found in a variety of sources (Brown, 1986; Azoury & Warmuth, 2001; Banerjee et al., 2005; Wainwright & Jordan, 2008); further bibliographic remarks will appear throughout, and in Appendix G. Finally, parallel to the completion of this manuscript, another group has developed exponential families under similarly relaxed conditions, but from the perspective of maximum entropy and convex duality (Csiszár & Matúš, 2012).

### 1.4. Notation

The following concepts from convex analysis are used throughout the text; the interested reader is directed to the seminal text of Rockafellar (1970). A set is convex when the line segment between any two of its elements is again within the set. The epigraph of a function $f : \mathbb{R}^n \to \bar{\mathbb{R}}$, where $\bar{\mathbb{R}} = \mathbb{R} \cup \{\pm\infty\}$, is the set of points bounded below by $f$; i.e., the set $\{(x, r) : x \in \mathbb{R}^n, r \geq f(x)\} \subseteq \mathbb{R}^n \times \bar{\mathbb{R}}$. A function is convex when its epigraph is convex, and closed when its epigraph is closed. The domain $\text{dom}(f)$ of a function $f : \mathbb{R}^n \to \bar{\mathbb{R}}$ is the subset of inputs not mapping to $+\infty$: $\text{dom}(f) = \{x \in \mathbb{R}^n : f(x) < \infty\}$. A function is proper if $\text{dom}(f)$ is nonempty, and $f$ never takes on the value $-\infty$. The Bregman divergences in this manuscript will be generated from closed proper convex functions.

The conjugate of a function $f$ is the function $f^*(\phi) := \sup_x \langle \phi, x \rangle - f(x)$; when $f$ is closed proper convex, so is $f^*$, and moreover $f^{**} = f$. A subgradient $g$ to a function $f$ at $y \in \text{dom}(f)$ provides an affine lower bound: for any $x \in \mathbb{R}^n$, $f(x) \geq f(y) + \langle g, x - y \rangle$. The set of all subgradients at a point $y$ is denoted by $\partial f(y)$ (which is easily empty). The directional derivative $f'(y; d)$ of a function $f$ at $y$ in direction $d$ is $\lim_{t \downarrow 0} (f(y + td) - f(y))/t$.

The affine hull of a set $S \subseteq \mathbb{R}^n$ is the smallest affine set containing it. If $S$ is translated and rotated so that its affine hull is some $\mathbb{R}^d \subseteq \mathbb{R}^n$, then its interior within $\mathbb{R}^d$ is its relative interior within $\mathbb{R}^n$. Said another way, the relative interior $\text{ri}(S)$ is the interior of $S$ with respect to the $\mathbb{R}^n$ topology relativized to the affine hull of $S$. Although functions in this manuscript will generally be closed, their domains are often (relatively) open.

Convex functions will be defined over $\mathbb{R}^n$, but it will be useful to treat data as lying in an abstract space $\mathcal{X}$, and a statistic map $\tau : \mathcal{X} \to \mathbb{R}^n$ will embed examples in the desired Euclidean space. This map, which will also be overloaded to handle finite subsets of $\mathcal{X}$, will eventually incorporate the smoothing procedure.

The cluster cost will be denoted by $\phi$, or $\phi_{f,\tau}$ to make the underlying convex function and statistic map clear; similarly, the merge cost is denoted by $\Delta$ and $\Delta_{f,\tau}$.

## 2. Bregman divergences

Given a convex function $f : \mathbb{R}^n \to \bar{\mathbb{R}}$, the Bregman divergence $\mathsf{B}_f(\cdot, y)$ is the gap between $f$ and its linearization at $y$. Typically, $f$ is differentiable, and so $\mathsf{B}_f(x, y) = f(x) - f(y) - \langle \nabla f(y), x - y \rangle$.

**Definition 2.1.** Given a convex function $f : \mathbb{R}^n \to \bar{\mathbb{R}}$, the corresponding Bregman divergence between $x, y \in \text{dom}(f)$ is

$$\mathsf{B}_f(x, y) := f(x) - f(y) + f'(y; y - x). \qquad \Diamond$$

Unlike gradients, directional derivatives are well-defined whenever a convex function is finite, although they can be infinite on the relative boundary of $\text{dom}(f)$ (Rockafellar, 1970, Theorems 23.1, 23.3, 23.4).

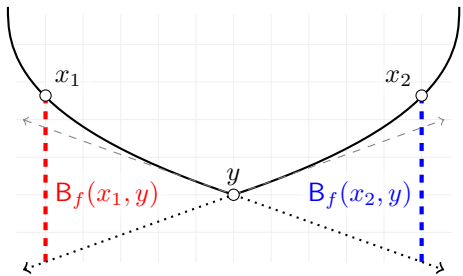

Figure 1. Bregman divergences with respect to a reference point $y$ at which $f$ is nondifferentiable. The thick (red or blue) dashed lines denote the divergence values themselves; they travel down from $f$ to the negated sublinear function $x \mapsto f(y) - f'(y; y - x)$, here a pair of dotted rays. Noting Proposition 2.3 and fixing some $x_i$, the subgradient at $y$ farthest from $x_i$ is one of these dotted rays together with its dashed, gray extension. The gray extensions, taken together, represent the sublinear function $x \mapsto f(y) + f'(y; x - y)$.

Noting that $f'(y; y - x) \geq -f'(y; x - y)$ (Rockafellar, 1970, Theorem 23.1), it may seem closer to the original expression to instead use $f(x) - f(y) - f'(y; x - y)$ (which is thus bounded above by $\mathsf{B}_f(x, y)$); however, it will later be shown that $\mathsf{B}_f(\cdot, y)$ is convex, which fails if the directional derivative is flipped. This distinction is depicted in Figure 1.

**Example 2.2.** In the case of the differentiable convex function $f_2 = \|\cdot\|_2^2$, $\mathsf{B}_{f_2}(x, y) = \|x - y\|_2^2$ follows by noting $f_2'(y; y - x) = \langle 2y, y - x \rangle$. To analyze the case of $f_1 = \|\cdot\|_1$, first consider the univariate case $h = |\cdot|$. Either by drawing a picture or checking $h'(\cdot; \cdot)$ from definition, it follows that

$$\mathsf{B}_h(x, y) := \begin{cases} 0 & \text{when } xy > 0, \\ 2|x| & \text{otherwise.} \end{cases}$$

Then noting that $f_1'(\cdot; \cdot)$ decomposes coordinate-wise, it follows that $\mathsf{B}_{f_1}(x, y) = \sum_i \mathsf{B}_h(x_i, y_i)$. Said another way, $\mathsf{B}_{f_1}$ is twice the $l^1$ distance from $x$ to the farthest orthant containing $y$, which bears a resemblance to the hinge loss. ◇

$\mathsf{B}_f$ can also be written in terms of subgradients.

**Proposition 2.3.** *Let a proper convex $f$ and $y \in \mathrm{ri}(\mathrm{dom}(f))$ be given. Then for any $x \in \mathrm{dom}(f)$:*

- *$f'(y; y - x)$ and $\mathsf{B}_f(x, y)$ are finite, and*

- *$\mathsf{B}_f(x, y) := \max_{g \in \partial f(y)} f(x) - f(y) - \langle g, x - y \rangle$.*

The above characterization will be extremely useful in proofs, where the existence of a maximizing subgradient $\bar{g}_{y,x}$ will frequently be invoked.

**Remark 2.4.** Given $x \in \mathrm{dom}(f)$ and a dual element $g \in \mathbb{R}^n$, another nondifferentiable generalization of Bregman divergence, due to Gordon (1999), is

$$\mathsf{D}_f(x, g) := f(x) + f^*(g) - \langle g, x \rangle.$$

Now suppose there exists $y \in \mathrm{ri}(\mathrm{dom}(f))$ with $g \in \partial f(y)$; the Fenchel-Young inequality (Rockafellar, 1970, Theorem 23.5) grants $\mathsf{D}_f(x, g) = f(x) - f(y) - \langle g, x - y \rangle$. Thus, by Proposition 2.3,

$$\mathsf{B}_f(x, y) := \max\{\mathsf{D}_f(x, g) : g \in \partial f(y)\}. \quad \diamond$$

To sanity check $\mathsf{B}_f$, Appendix A states and proves a number of key Bregman divergence properties, generalized to the case of nondifferentiability. The following list summarizes these properties; in general, $f$ is closed proper convex, $y \in \mathrm{ri}(\mathrm{dom}(f))$, and $x \in \mathrm{dom}(f)$.

- $\mathsf{B}_f(\cdot, y)$ is convex, proper, nonnegative, and $\mathsf{B}_f(y, y) = 0$.

- When $f$ is strictly convex, $\mathsf{B}_f(x, y) = 0$ iff $x = y$.

- Given $g_x \in \mathrm{ri}(\mathrm{dom}(f^*))$, $\sup_{x \in \partial f^*(g_x)} \mathsf{B}_f(x, y) = \sup_{g_y \in \partial f(y)} \mathsf{B}_{f^*}(g_y, g_x)$.

- Under some regularity conditions on $f$, a generalization of the Pythagorean theorem holds, with $\mathsf{B}_f$ replacing squared Euclidean distance.

Over and over, this section depends on relative interiors. What's so bad about the relative boundary? The directional derivatives and subgradients break down. If $y \in \mathrm{relbd}(\mathrm{dom}(f))$ and $x \in \mathrm{ri}(\mathrm{dom}(f))$, then $f'(y; y - x) = \infty = \mathsf{B}_f(x, y)$, and there exists no maximizing subgradient as in Proposition 2.3; in fact, one can not in general guarantee the existence of any subgradients at all.

In just a moment, the cluster model will be developed, where it will be very easy for the second argument argument of $\mathsf{B}_f$ to lie on $\mathrm{relbd}(\mathrm{dom}(f))$, rendering the divergences infinite and cluster costs meaningless. Worse, it is frequently the case $\mathrm{dom}(f)$ is relatively open, meaning the relative boundary is not in $\mathrm{dom}(f)$! The smoothing methods of Section 5 work around these issues. Their approach is simple enough: they just push relative boundary points inside the relative interior.

## 3. Cluster model

Let a finite collection $C$ of points $\{x_i\}_{i=1}^m$ in some abstract space $\mathcal{X}$ — say, documents or vectors — be given. In order to cluster these with Bregman divergences, the first step is to map them into $\mathbb{R}^n$.

**Definition 3.1.** A statistic map $\tau$ is any function from $\mathcal{X}$ to $\mathbb{R}^n$. Given a finite set $C \subseteq \mathcal{X}$, overload $\tau$ via averages: $\tau(C) := |C|^{-1} \sum_{x \in C} \tau(x)$. ◇

For now, it suffices to think of $\tau$ as the identity map (with $\mathcal{X} = \mathbb{R}^n$), with an added convenience of computing means. Section 4, however, will rely on $\tau$ when constructing exponential families.

**Definition 3.2.** Given a statistic map $\tau : \mathcal{X} \to \mathbb{R}^n$ and convex function $f$, the cost of a single cluster $C$ is

$$\phi_{f,\tau}(C) := \sum_{x \in C} \mathsf{B}_f(\tau(x), \tau(C)).$$

(This cost was the basis for *Bregman hard clustering* (Banerjee et al., 2005).) ◇

**Example 3.3** ($k$-means cost). Choose $\mathcal{X} = \mathbb{R}^n$, $\tau(x) = x$, and $f = \|\cdot\|_2^2$. As discussed in Example 2.2, $\mathsf{B}_f(x,y) = \|x-y\|_2^2$, and so $\phi_{f,\tau}(C) = \sum_{x \in C} \|x - \tau(C)\|_2^2$, precisely the $k$-means cost. ◇

As such, $\tau(C)$ plays the role of a cluster center. While this may be intuitive for the $k$-means cost, it requires justification for general Bregman divergences. The following definition and results bridge this gap.

**Definition 3.4.** A convex function $f$ is *relatively (Gâteaux) differentiable* if, for any $y \in \mathrm{ri}(\mathrm{dom}(f))$, there exists $g$ (necessarily any subgradient) so that, for any $x \in \mathrm{dom}(f)$, $f'(y; y-x) = \langle g, y-x \rangle$. ◇

Every differentiable function is relatively differentiable (with $g = \nabla f(y)$); fortunately, many relevant convex functions, in particular those used to construct Bregman divergences between exponential family distributions (cf. Proposition 4.5), will be relatively differentiable.

Under relative differentiability, Bregman divergences admit a bias-variance style decomposition, which immediately justifies the choice of centroid $\tau(C)$.

**Lemma 3.5.** *Let a proper convex relatively differentiable $f$, points $\{x_i\}_{i=1}^m \subset \mathbb{R}^n$, and weights $\{\alpha_i\}_{i=1}^m \subset \mathbb{R}$ be given, with $\mu := \sum_i \alpha_i x_i / (\sum_j \alpha_j) \in \mathrm{ri}(\mathrm{dom}(f))$. Then, given any point $y \in \mathrm{ri}(\mathrm{dom}(f))$,*

$$\sum_{i=1}^m \alpha_i \mathsf{B}_f(x_i, y) = \sum_{i=1}^m \alpha_i \mathsf{B}_f(x_i, \mu) + \left(\sum_{i=1}^m \alpha_i\right) \mathsf{B}_f(\mu, y).$$

**Corollary 3.6.** *Suppose the convex function $f$ is relatively differentiable, let any statistic map $\tau$ and any finite cluster $C \subseteq \mathcal{X}$ be given. Then $\phi_{f,\tau}(C) = \inf_{y \in \mathbb{R}^n} \sum_{x \in C} \mathsf{B}_f(\tau(x), y)$.*

*Proof.* Use $\mu := \tau(C) = |C|^{-1} \sum_{x \in C} \tau(x)$, Lemma 3.5, and $\mathsf{B}_f \geq 0$. □

Continuing, the stage is set to finally construct the Bregman merge cost.

**Definition 3.7.** Given two finite subsets $C_1, C_2$ of $\mathcal{X}$, the cluster merge cost is simply growth in total cost:

$$\Delta_{f,\tau}(C_1, C_2) = \phi_{f,\tau}(C_1 \cup C_2) - \sum_{j \in \{1,2\}} \phi_{f,\tau}(C_j). \quad \diamond$$

The above expression seems to imply that the computational cost of $\Delta$ scales with the number of points. But in fact, one need only look at the relevant centers.

**Proposition 3.8.** *Let a proper convex relatively differentiable $f$ and two finite subsets $C_1, C_2$ of $\mathcal{X}$ with $\tau(C_i) \in \mathrm{ri}(\mathrm{dom}(f))$ be given. Then*

$$\Delta_{f,\tau}(C_1, C_2) = \sum_{j \in \{1,2\}} |C_j| \mathsf{B}_f(\tau(C_j), \tau(C_1 \cup C_2)).$$

**Example 3.9** (Ward/$k$-means merge cost). Continuing with the $k$-means cost from Example 3.3, note that for $j \in \{1, 2\}$,

$$\|\tau(C_j) - \tau(C_1 \cup C_2)\|_2 = \frac{|C_{3-j}| \cdot \|\tau(C_1) - \tau(C_2)\|_2}{|C_1| + |C_2|}.$$

Plugging this into the simplification of $\Delta_{f,\tau}$ provided by Proposition 3.8,

$$\Delta_{f,\tau}(C_1, C_2) = \sum_{j \in \{1,2\}} \frac{|C_j||C_{3-j}|^2}{(|C_1| + |C_2|)^2} \|\tau(C_1) - \tau(C_2)\|_2^2$$
$$= \frac{|C_1||C_2|}{|C_1| + |C_2|} \|\tau(C_1) - \tau(C_2)\|_2^2.$$

This is exactly the Ward merge cost. ◇

## 4. Exponential families

So far, this manuscript has developed a mathematical basis to clustering with Bregman divergences. But what does it matter, if examples of meaningful Bregman divergences are few and far between?

The primary mechanism for constructing meaningful merge costs is to model the clusters as exponential family distributions. Throughout this section, let $\nu$ be any measure over $\mathcal{X}$, and further stipulate the statistic map $\tau$ is $\nu$-measurable.

**Definition 4.1.** Given a measurable statistic map $\tau$ and a vector $\theta \in \mathbb{R}^n$ of *canonical parameters*, the corresponding exponential family distribution has density

$$p_\theta(x) := \exp(\langle \tau(x), \theta \rangle - \psi(\theta)),$$

where the normalization term $\psi$, typically called the *cumulant* or *log partition function*, is simply

$$\psi(\theta) = \ln \int \exp(\langle \tau(x), \theta \rangle) d\nu(x). \quad \diamond$$

Many standard distributions have this representation.

**Example 4.2.** Choose $\mathcal{X} = \mathbb{R}^d$ with $\nu$ being Lebesgue measure, and $n = d(d+1)$, i.e. $\mathbb{R}^n = \mathbb{R}^{d(d+1)}$. The map $\tau(x) = (x, xx^\top)$ will provide for Gaussian densities. In particular, starting from the familiar form, with mean $\mu \in \mathbb{R}^d$ and positive definite covariance $\Sigma \in \mathbb{R}^{d^2}$, the density at $x$, $p_\theta(x)$, is

$$\frac{\exp(-(x-\mu)^\top \Sigma^{-1}(x-\mu)/2)}{\sqrt{(2\pi)^d |\Sigma|}}$$

$$= \exp\Big(\langle \tau(x), (\Sigma^{-1}\mu, -\Sigma^{-1}/2)\rangle$$
$$- \frac{1}{2}\ln((2\pi)^d |\Sigma| \exp(\mu^\top \Sigma^{-1}\mu))\Big).$$

In other words, $\theta = (\Sigma^{-1}\mu, -\Sigma^{-1}/2)$. Notice that $\psi$ (here expanded as $\frac{1}{2}\ln(\ldots)$) and $\theta$ do not make sense if $\Sigma$ is merely positive semi-definite. $\diamond$

So far so good, but where's the convex function, and does the definition of $p_\theta$ even make sense?

**Proposition 4.3.** *Given a measurable statistic map $\tau$, the function $\psi$ is well-defined, closed, convex, and never takes on the value $-\infty$.*

**Remark 4.4.** Notice that Proposition 4.3 did not provide that $\psi$ is proper, only that it is never $-\infty$. Unfortunately, more can not be guaranteed: if $\nu$ is Lebesgue measure over $\mathbb{R}$ and $\tau(x) = 0$ for all $x$, then every parameter choice $\theta \in \mathbb{R}$ has $\psi(\theta) = \infty$. It is therefore necessary to check, for any provided $\tau$, whether $\text{dom}(\psi)$ is nonempty. $\diamond$

Not only is $\psi$ closed convex, it is about as well-behaved as any function discussed in this manuscript.

**Proposition 4.5.** *Suppose $\text{dom}(\psi)$ is nonempty. Then $\psi$ is relatively differentiable; in fact, given any $\theta \in \text{ri}(\text{dom}(\psi))$, any $\hat\tau \in \partial\psi(\theta)$, and any $\xi \in \text{dom}(\psi)$,*

$$\psi'(\theta; \xi - \theta) = \langle \hat\tau, \xi - \theta\rangle = \int \langle \tau(x), \xi - \theta\rangle p_\theta(x) d\nu(x).$$

If $\psi$ is fully differentiable at $\theta$, then $\nabla\psi(\theta) = \int \tau p_\theta$. Since $\psi$ is closed, given $\hat\tau \in \partial\psi(\theta)$, it follows that $\theta \in \partial\psi^*(\hat\tau)$. There is still cause for concern that other subgradients at $\hat\tau$ lead to different densities, but as will be shown below, this does not happen.

Now that a relevant convex function $\psi$ has been identified, the question is whether $\mathsf{B}_\psi$ (or $\mathsf{B}_{\psi^*}$) provide a reasonable notion of distance amongst densities.

This will be answered in two ways. To start, recall the Kullback-Leibler divergence $\mathsf{K}$ between densities $p, q$:

$$\mathsf{K}(p, q) = \int p(x) \ln\left(\frac{p(x)}{q(x)}\right) d\nu(x).$$

**Theorem 4.6.** *Let any $\theta_1, \theta_2 \in \text{ri}(\text{dom}(\psi))$ and any $\hat\tau_1 \in \partial\psi(\theta_1), \hat\tau_2 \in \partial\psi(\theta_2)$ be given, where $\partial\psi^*(\hat\tau_2) \subseteq \text{ri}(\text{dom}(\psi))$ (for instance, if $\text{dom}(\psi)$ is relatively open). Then*

$$\mathsf{K}(p_{\theta_1}, p_{\theta_2}) = \mathsf{B}_\psi(\theta_2, \theta_1) = \mathsf{B}_{\psi^*}(\hat\tau_1, \hat\tau_2).$$

*Furthermore, if $\theta_1 \in \partial\psi^*(\hat\tau_2)$, then $p_{\theta_1} = p_{\theta_2}$ $\nu$-a.e..*

Motivated by Proposition 4.5 and Theorem 4.6, the choice here is to base the cluster model on $\mathsf{B}_{\psi^*}$.

Given two clusters $\{C_i\}_{i=1}^2$, set $\hat\tau_i := \tau(C_i)$. When working with these clusters, it is entirely sufficient to store only these statistics and the cluster sizes, since $\tau(C_1 \cup C_2) = |C_1 \cup C_2|^{-1}(|C_1|\hat\tau_1 + |C_2|\hat\tau_2)$. Assuming for interpretability that $\psi$ is differentiable, since $\psi$ is closed, $\psi^{**} = \psi$, and thus $\nabla\psi(\theta_1) = \int \tau p_{\theta_1} = \hat\tau_1$; that is to say, these distributions have their (aptly named) *mean parameterizations* as their means. And as provided by Theorem 4.6, even if differentiability fails, various subgradients of the same mean all effectively represent the same distributions.

**Example 4.7.** Suppose $\mathcal{X}$ is a finite set, representing a vocabulary with $n$ words, and $\nu$ is counting measure over $\mathcal{X}$. The statistic map $\tau$ converts word $k$ into the $k^{\text{th}}$ basis vector $\mathbf{e}_k$. Let $\hat\tau \in \mathbb{R}^n_{++}$ represent the mean parameters of a multinomial over this set; observe that

$$p_\theta(\mathbf{e}_i) = \langle \tau(i), \hat\tau\rangle$$
$$= \exp(\langle \mathbf{e}_i, \ln\hat\tau\rangle) - \ln\int \exp(\langle \tau(k), \ln\hat\tau\rangle) d\nu(k).$$

That is to say, the canonical parameter vector is $\theta = \ln\hat\tau$, the coordinate-wise logarithm of the mean parameters. Proposition 4.5 can be verified directly: $(\nabla\psi(\theta))_i = e^{\theta_i}/\sum_k e^{\theta_k} = \hat\tau$. Similarly, given another multinomial with mean parameters $\hat\tau' \in \mathbb{R}^n_{++}$ and canonical parameters $\theta' = \ln\hat\tau'$,

$$\mathsf{K}(p_\theta, p_{\theta'}) = \sum_{i=1}^n \hat\tau_i \ln\left(\frac{\hat\tau_i}{\hat\tau'_i}\right).$$

The notation $\mathbb{R}^n_{++}$ means strictly positive coordinates: no word can have zero probability. Without this restriction, it is not possible to map into the canonical parameter space. This is precisely the scenario the smoothing methods of Section 5 will work around: the provided clusters are on the relative boundary of $\text{dom}(\psi^*)$, which is either not part of $\text{dom}(\psi^*)$ at all, or as is the case here, causes degenerate Bregman divergences (infinite valued, and lacking subgradients). $\diamond$

**Remark 4.8.** The multinomials in Example 4.7 have an overcomplete representation: scaling any canonical parameter vector by a constant gives the same mean parameter. In general, if two relative interior canonical parameters $\theta_1 \neq \theta_2$ have a common subgradient $\hat{\tau} \in \partial\psi(\theta_1) \cap \partial\psi(\theta_2)$, then it follows that $\{\theta_1, \theta_2\} \subset \partial\psi^*(\hat{\tau})$ (Rockafellar, 1970, Theorem 23.5). That is to say: this scenario leads to mean parameters which have distinct subgradients, and are thus points of nondifferentiability within $\mathrm{ri}(\mathrm{dom}(\psi^*))$, which necessitate the generalized development of Bregman divergences in this manuscript. ◊

A further example of Gaussians appears in Appendix C.

The second motivation for $\Delta_{\psi^*,\tau}$ is an interpretation in terms of model fit.

**Theorem 4.9.** *Fix some measurable statistic map $\tau$, and let two finite subsets $C_1, C_2$ of $\mathcal{X}$ be given with mean parameters $\{\tau(C_1), \tau(C_2)\} = \{\hat{\tau}_1, \hat{\tau}_2\} \subseteq \mathrm{ri}(\mathrm{dom}(\psi^*))$. Choose any canonical parameters $\theta_i \in \partial\psi^*(\hat{\tau}_i)$, and for convenience set $C_3 := C_1 \cup C_2$, with mean parameter $\hat{\tau}_3$ and any canonical parameter $\theta_3 \in \partial\psi^*(\hat{\tau}_3)$. Then*

$$\Delta_{\psi^*,\tau}(C_1, C_2) = \sum_{i \in \{1,2\}} \sum_{x \in C_i} \ln p_{\theta_i}(x) - \sum_{x \in C_3} \ln p_{\theta_3}(x).$$

## 5. Smoothing

The final piece of the technical puzzle is the smoothing procedure: most of the above properties — for instance, that $\mathsf{B}_f(\tau(C_1), \tau(C_2)) < \infty$ — depend on $\tau(C_2) \in \mathrm{ri}(\mathrm{dom}(f))$. Relative boundary points lead to degeneracies; for example, this characterizes the Gaussian degeneracy identified in the introduction.

**Definition 5.1.** Given a (nonempty) convex set $S$, a statistic map $\tau : \mathcal{X} \to \mathbb{R}^n$ is a *smoothing statistic map* for $S$ if, given any finite set $C \subseteq \mathcal{X}$, $\tau(C) \in \mathrm{ri}(S)$. ◊

It turns out to be very easy to construct smoothing statistic maps.

**Theorem 5.2.** *Let a nonempty convex set $S$ be given. Let $\tau_0 : \mathcal{X} \to \mathbb{R}^n$ be a statistic map satisfying, for any finite $C \subseteq \mathcal{X}$, $\tau_0(C) \in \mathrm{cl}(S)$. Let $z \in \mathrm{ri}(S)$ and $\alpha \in (0,1)$ be arbitrary. Given any finite set $C \subseteq \mathcal{X}$, define the maps*

$$\tau_1(C) := (1-\alpha)\tau_0(C) + \alpha z,$$
$$\tau_2(C) := \tau_0(C) + \alpha z.$$

*In general, $\tau_1$ is a smoothing statistic map for $S$. If additionally $S$ is a convex cone, then $\tau_2$ is also a smoothing statistic map for $S$.*

The following two examples smooth Gaussians and multinomials via Theorem 5.2. The parameters $\alpha$ and $z$ are chosen from data, and moreover satisfy $\|\alpha z\| \downarrow 0$ as the total amount of available data grows; that is to say, $\tau$ will more and more closely match $\tau_0$.

**Example 5.3** (Smoothing multinomials.)**.** The mean parameters to a multinomial lie within the probability simplex, a compact convex set. As discussed in Example 4.7, only the relative interior of the simplex provides canonical parameters. According to Theorem 5.2, all that remains in fixing this problem is to determine $\alpha z$.

The approach here is to interpret the provided multinomial $\tau_0(C) = \hat{\tau}$ as based on a finite sample of size $m$, and thus the true parameters lie within some confidence interval around $\hat{\tau}$; crucially, this confidence interval intersects the relative interior of the probability simplex. One choice is a Bernstein-based upper confidence estimate $\tau(C) = \tau_0(C) + \mathcal{O}(1/m + \sqrt{p(1-p)/m})$, where $p = 1/n$. ◊

**Example 5.4** (Smoothing Gaussians.)**.** In the case of Gaussians, as discussed in Example 4.2, only positive definite covariance matrices are allowed. But this set is a convex cone, so Theorem 5.2 reduces the problem to finding a sensible element to add in.

Consider the case of singleton clusters. Adding a full-rank covariance matrix in to the observed zero covariance matrix is like replacing this singleton with a constellation of points centered at it. Equivalently, each point is replaced with a tiny Gaussian, which is reminiscent of nonparametric density estimation techniques. Therefore one option is to use bandwidth selection techniques; the experiments of Section 7 use the "normal reference rule" (Bowman & Azzalini, 1997, Section 2.4.2), trying both the approach of estimating a bandwidth for each coordinate (suffix `-nd`), and computing one bandwidth for every direction uniformly, and simply adding a rescaling of the identity matrix to the sample covariance (suffix `-n`). ◊

When there is a probabilistic interpretation of the clusters, and in particular $\tau(C)$ may be viewed as a maximum likelihood estimate, another approach is to choose some prior over the parameters, and have $\tau$ produce a MAP estimate which also lies in the relative interior. As stated, this approach will differ slightly from the one presented here: the weight on the added element will scale with the cluster size, rather than the size of the full data, and the relationship of $\tau(C_1 \cup C_2)$ to $\tau(C_1)$ and $\tau(C_2)$ becomes less clear.

## 6. Clustering algorithm

The algorithm appears in Algorithm 1. Letting $T_{\Delta_f, \tau}$ denote an upper bound on the time to calculate a

**Algorithm 1** AGGLOMERATE.
**Input** Merge cost $\Delta_{f,\tau}$, points $\{x_i\}_{i=1}^m \subseteq \mathcal{X}$.
**Output** Hierarchical clustering.

  Initialize forest as $\mathscr{F} := \{\{x_i\} : i \in [m]\}$.
  **while** $|\mathscr{F}| > 1$ **do**
    Let $\{C_i, C_j\} \subseteq \mathscr{F}$ be any pair minimizing $\Delta_{f,\tau}(C_i, C_j)$, as computed by Proposition 3.8.
    Remove $\{C_i, C_j\}$ from $\mathscr{F}$, add in $C_i \cup C_j$.
  **end while**
  **return** the single tree within $\mathscr{F}$.

Table 1. Dendrogram purity on Euclidean and text data.

|  | c-link | s-link | km | dg-nd | g-n |
|---|---|---|---|---|---|
| glass | 0.46 | 0.45 | 0.50 | 0.49 | **0.54** |
| spam | 0.59 | 0.58 | 0.59 | **0.65** | 0.60 |
| mnist35 | 0.59 | 0.51 | 0.69 | 0.62 | **0.73** |

|  | c-link | s-link | multi |
|---|---|---|---|
| 20n-e | 0.60 | 0.50 | **0.93** |
| 20n-h | 0.54 | 0.52 | **0.56** |
| 20n-b | 0.31 | 0.29 | **0.62** |

single merge cost, a brute-force implementation (over $m$ points) takes space $\mathcal{O}(m)$ and time $\mathcal{O}(m^3 T_{\Delta_{f,\tau}})$, whereas caching merge cost computations in a min-heap requires space $\mathcal{O}(m^2)$ and time $\mathcal{O}(m^2(\lg(m) + T_{\Delta_{f,\tau}}))$. Please refer Appendix E for more notes on running times, and a depiction of sample hierarchies over synthetic data.

If Proposition 3.8 is used to compute $\Delta_{f,\tau}$, then only the sizes and means of clusters need be stored, and computing this merge cost involves just two Bregman divergences calculations. As the new mean is a convex combination of the two old means, computing it takes time $\mathcal{O}(n)$. The Bregman cost itself can be more expensive; for instance, as discussed with Gaussians in Appendix C, it is necessary to invert a matrix, meaning $\mathcal{O}(n^3)$ steps.

## 7. Empirical results

Trees generated by AGGLOMERATE are evaluated in two ways. First, cluster compatibility scores are computed via dendrogram purity and initialization quality for EM upon mixtures of Gaussians. Secondly, cluster features are fed into supervised learners.

There are two kinds of data: Euclidean (points in some $\mathbb{R}^n$), and text data. There are three Euclidean data sets: UCI's `glass` (214 points in $\mathbb{R}^9$); 3s and 5s from the `mnist` digit recognition problem (1984 training digits and 1984 testing digits in $\mathbb{R}^{49}$, scaled down from the original 28x28); UCI's `spambase` data (2301 train and 2300 test points in $\mathbb{R}^{57}$). Text data is drawn from the 20 newsgroups data, which has a vocabulary of 61,188 words; a difficult dichotomy (`20n-h`), the pair `alt.atheism/talk.religion.misc` (856 train and 569 test documents); an easy dichotomy (`20n-e`), the pair `rec.sport.hockey/sci.crypt` (1192 train and 794 test documents). Finally, `20n-b` collects these four groups into one corpus.

The various trees are labeled as follows. `s-link` and `c-link` denote single and complete linkage, where $l^1$ distance is used for text, and $l^2$ distance is used for Euclidean data. `km` is the Ward/$k$-means merge cost. `g-n` fits full covariance Gaussians, whereas `dg-nd` fits diagonal covariance Gaussians; smoothing follows the data-dependent scheme of Example 5.4. `multi` fits multinomials, with the smoothing scheme of Example 5.3.

### 7.1. Cluster compatibility

Table 1 contains cluster purity scores, a standard dendrogram quality measure, defined as follows. For any two points with the same label $l$, find the smallest cluster $C$ in the tree which contains them both; the purity with respect to these two points is the fraction of $C$ having label $l$. The purity of the dendrogram is the weighted sum, over all pairs of points sharing labels, of pairwise purities. The `glass`, `spam`, and `20newsgroups` data appear in Heller & Ghahramani (2005); although a direct comparison is difficult, since those experiments used subsampling and randomized purity, the Euclidean experiments perform similarly, and the text experiments fare slightly better here.

For another experiment, now assessing the viability of AGGLOMERATE as an initialization to EM applied to mixtures of Gaussians, please see Appendix F.

### 7.2. Feature generation

The final experiment is to use dendrograms, built from training data, to generate features for classification tasks. Given a budget of features $k$, the top $k$ clusters $\{C_i\}_{i=1}^k$ of a specified dendrogram are chosen, and for any example $x$, the $i^{\text{th}}$ feature is $\Delta(C_i, \{x\})$. Statistically, this feature measures the amount by which the model likelihood degrades when $C_i$ is adjusted to accommodate $x$. The choice of tree was based on training set purity from Table 1. In all tests, the original features are discarded (i.e., only the $k$ generated features are used).

Figure 2 shows the performance of logistic regression classifiers using tree features, as well as SVD features. The stopping rule used validation set performance.

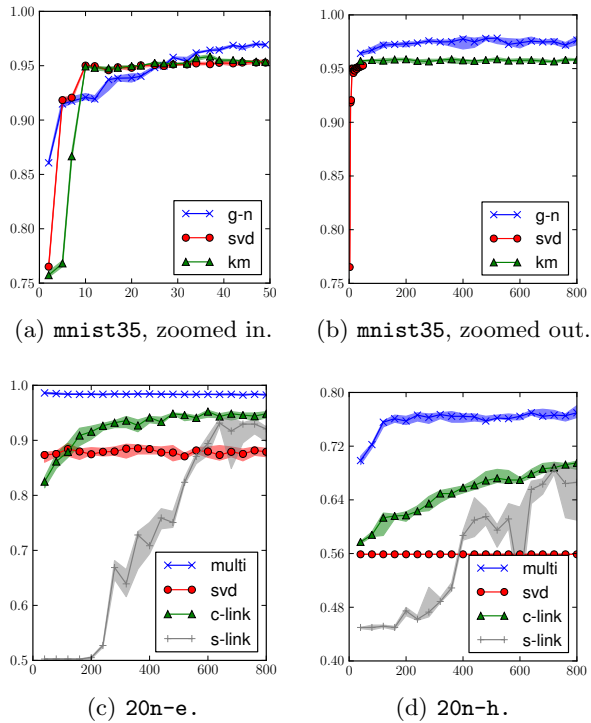

Figure 2. Comparison of dendrogram features to SVD features; y-axis denotes classification accuracy on test data, x-axis denotes #features. In the first two plots, `mnist35` was used. The SVD can only produce as many features as the dimension of the data, but the proposed tree continues to improve performance beyond this point. For the text data tasks `20n-e` and `20n-h`, tree methods strongly outperform the SVD. Please see text for details.

## Acknowledgements

This work was graciously supported by the NSF under grant IIS-0713540.